%% file: main.tex
\documentclass[fleqn,10pt]{wlscirep}
\usepackage[utf8]{inputenc}
\usepackage[T1]{fontenc}
\usepackage{lineno}
\usepackage{subfigure}
\usepackage{soul}

\usepackage{hyperref}
\hypersetup{colorlinks=true}

\title{HIT-UAV: A high-altitude infrared thermal dataset for Unmanned Aerial Vehicle-based object detection\footnote{Scientific Data | (2023) https://doi.org/10.1038/s41597-023-02066-6}}

\author[1,2]{Jiashun Suo}
\author[3]{Tianyi Wang}
\author[3,*]{Xingzhou Zhang}
\author[1,2]{Haiyang Chen}
\author[1,2]{Wei Zhou}
\author[4]{Weisong Shi}
\affil[1]{Engineering Research Center of Cyberspace, Yunnan University, Kunming 650091, China}
\affil[2]{School of Software, Yunnan University, Kunming 650091, China}
\affil[3]{Research Center of Distributed Systems, Institute of Computing Technology, Chinese Academy of Sciences, Beijing 100190, China}
\affil[4]{Department of Computer and Information Sciences, University of Delaware, Newark, DE 19716, USA}

\affil[*]{corresponding author(s): Xingzhou Zhang (zhangxingzhou@ict.ac.cn)}

\begin{abstract}
\input{1_Abstract.tex}
\end{abstract}
\begin{document}

\flushbottom
\maketitle

\thispagestyle{empty}


\section*{Background \& Summary}
\input{2_Background.tex}


\section*{Methods}
\input{3_Method.tex}






 

\section*{Data Records}
\input{4_Data_records.tex}



\section*{Technical Validation}

\input{5_Technical_validation.tex}


\section*{Usage Notes}
\input{6_Uasge_notes.tex}



\section*{Code availability}
\input{7_code_availability.tex}



\input{8_Reference_bib.tex}



\section*{Acknowledgements}
This work was supported in part by the China Postdoctoral Science Foundation under Grant No.2021M693227, in part by the National Natural Science Foundation of China under Grant 62162067, 62101480, and 62072434, in part by the Yunnan Province Science Foundation under Grant No.202005AC160007, No.202001BB050076, Research and Application of Object Detection based on Artificial Intelligence. We are indebted to the anonymous referees for their constructive remarks, to the Cloud-Edge-Things Testbed of Nanjing Institute of Information Superbahn.

\section*{Author contributions}
Jiashun Suo designed the experiments, collected the data, annotated the images, and wrote the paper. 
Tianyi Wang annotated the images and coded the tools.
Xingzhou Zhang reviewed and edited the paper and supervised the work.
Haiyang Chen collected the data and annotated the images.
Wei Zhou reviewed the paper, supervised the work, and provided funding.
Weisong Shi reviewed and edited the paper.
All authors reviewed the manuscript.

\section*{Competing interests}
The authors declare no competing interests.

\end{document}

%% file: 1_Abstract.tex
We present the HIT-UAV dataset, a high-altitude infrared thermal dataset for object detection applications on Unmanned Aerial Vehicles (UAVs). 
The dataset comprises 2,898 infrared thermal images extracted from 43,470 frames in hundreds of videos captured by UAVs in various scenarios including schools, parking lots, roads, and playgrounds.
Moreover, the HIT-UAV provides essential flight data for each image, such as flight altitude, camera perspective, date, and daylight intensity.
For each image, we have manually annotated object instances with bounding boxes of two types (oriented and standard) to tackle the challenge of significant overlap of object instances in aerial images.
To the best of our knowledge, the HIT-UAV is the first publicly available high-altitude UAV-based infrared thermal dataset for detecting persons and vehicles.
We have trained and evaluated well-established object detection algorithms on the HIT-UAV.
Our results demonstrate that the detection algorithms perform exceptionally well on the HIT-UAV compared to visual light datasets since infrared thermal images do not contain significant irrelevant information about objects.
We believe that the HIT-UAV will contribute to various UAV-based applications and researches.
The dataset is freely available at \href{https://github.com/suojiashun/HIT-UAV-Infrared-Thermal-Dataset}{https://github.com/suojiashun/HIT-UAV-Infrared-Thermal-Dataset}.

%% file: 2_Background.tex
\begin{figure}[!ht]
\centering
\includegraphics[width=5.8in]{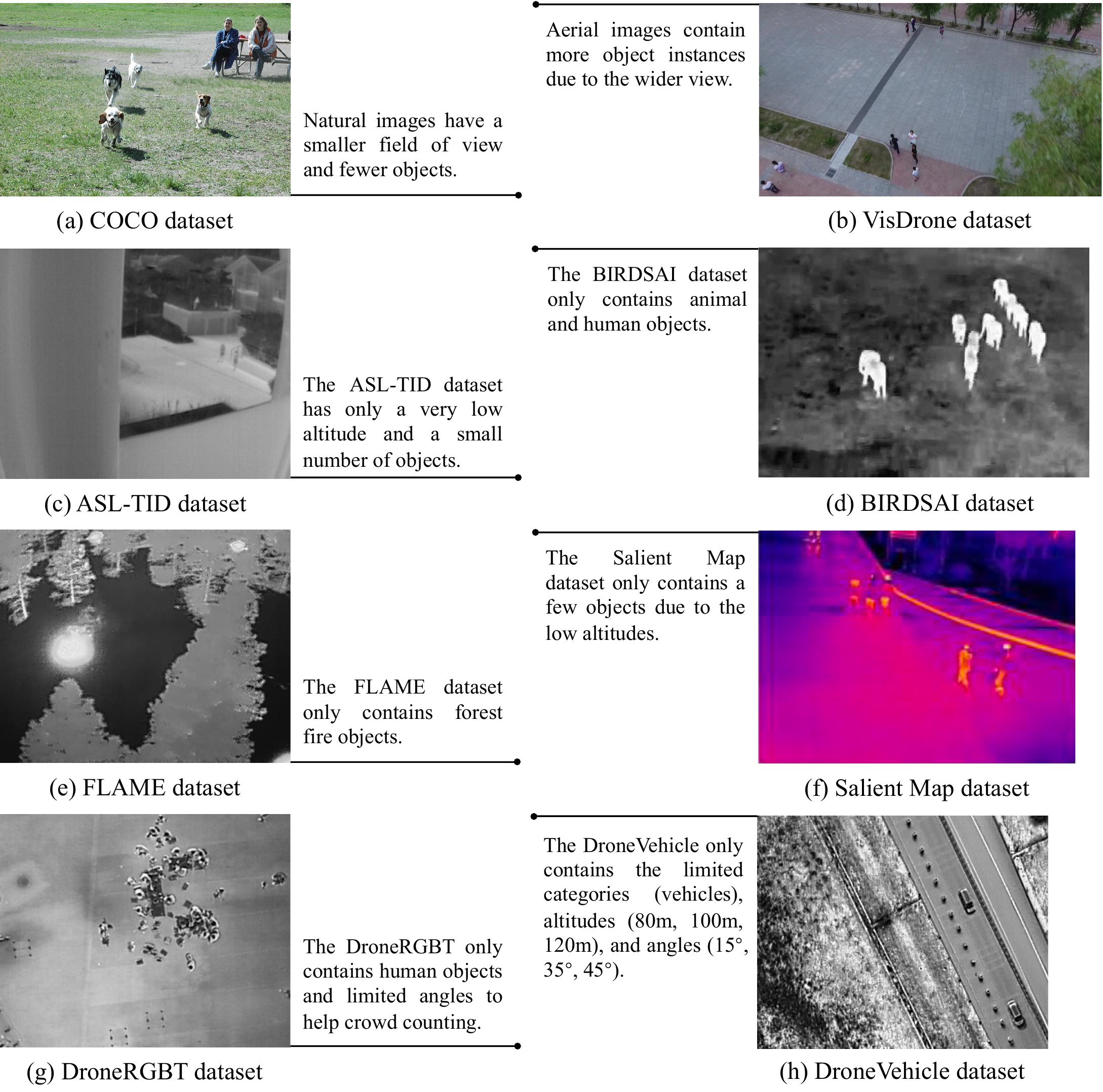}
\caption{The samples of different datasets.}
\label{table2_sample_different_dataset}
\end{figure}

Unmanned Aerial Vehicle (UAV)-based object detection algorithms are widely used for various domains such as forest inventory~\cite{wallace2012development}, mapping applications~\cite{samad2013potential}, traffic monitoring~\cite{heintz2007images}, and humanitarian relief~\cite{bravo2019use}.
With the rapid development of deep learning~\cite{pouyanfar2018survey} and edge computing~\cite{shi2016edge}, UAVs can now load edge computing devices to run artificial intelligence (AI) algorithms, thereby increasing their value in the aforementioned applications.
Motivated by the rapid development of object detection, several general datasets such as PASCAL VOC~\cite{everingham2010pascal}, MSCOCO~\cite{lin2014microsoft}, and ImageNet~\cite{deng2009imagenet} have been proposed to support algorithm training and evaluation.
However, unlike natural environments, aerial images contain more object instances due to the wider view, bringing more significant challenges.
Table \ref{table1_common_dataset_comparison} shows the average quantity of object bounding boxes per image for general datasets and the HIT-UAV~\cite{HIT-UAV}.
Compared to general datasets, the HIT-UAV~\cite{HIT-UAV} contains a higher average quantity of object bounding boxes.
Figure \ref{table2_sample_different_dataset} (a) and (b) use samples from the COCO and VisDrone datasets to show the differences between natural and aerial images.

Many datasets of aerial perspectives have been introduced to help improve the detection performance of algorithms.
The Stanford~\cite{robicquet2016learning}, UAV123~\cite{mueller2016benchmark}, CARPK~\cite{hsieh2017drone}, VisDrone~\cite{zhu2018vision}, and AU-AIR~\cite{bozcan2020air} datasets were introduced with visual light images.
The ASL-TID~\cite{portmann2014people}, BIRDSAI~\cite{bondi2020birdsai}, FLAME~\cite{shamsoshoara2021aerial}, DroneRGBT~\cite{dronergbt}, DroneVehicle~\cite{dronevehicle}, and Salient Map~\cite{li2020object} datasets were introduced with thermal infrared images.
The Salient Map dataset contains pedestrian and vehicle objects because the authors found there is no publicly available thermal dataset for detecting pedestrians and vehicles from the perspective of UAVs.

\begin{table}[!ht]
\centering
\caption{The average bounding box (Avg. Bbox) quantity of general datasets and the HIT-UAV.}
\scalebox{0.85}{
\begin{tabular}{c|c}
\toprule
\textbf{Dataset} & \textbf{Avg. Bbox} \\
\midrule
PASCAL VOC (2007 + 2012 version) & 2.89 \\
MSCOCO (2014 training + validation set) & 7.19 \\
ImageNet (2017 training set) & 1.37 \\
\midrule
\textbf{HIT-UAV} & \textbf{8.59}
\\
\bottomrule
\end{tabular}}
\label{table1_common_dataset_comparison}
\end{table}

\begin{figure}[!ht]
\centering
\includegraphics[width=4.6in]{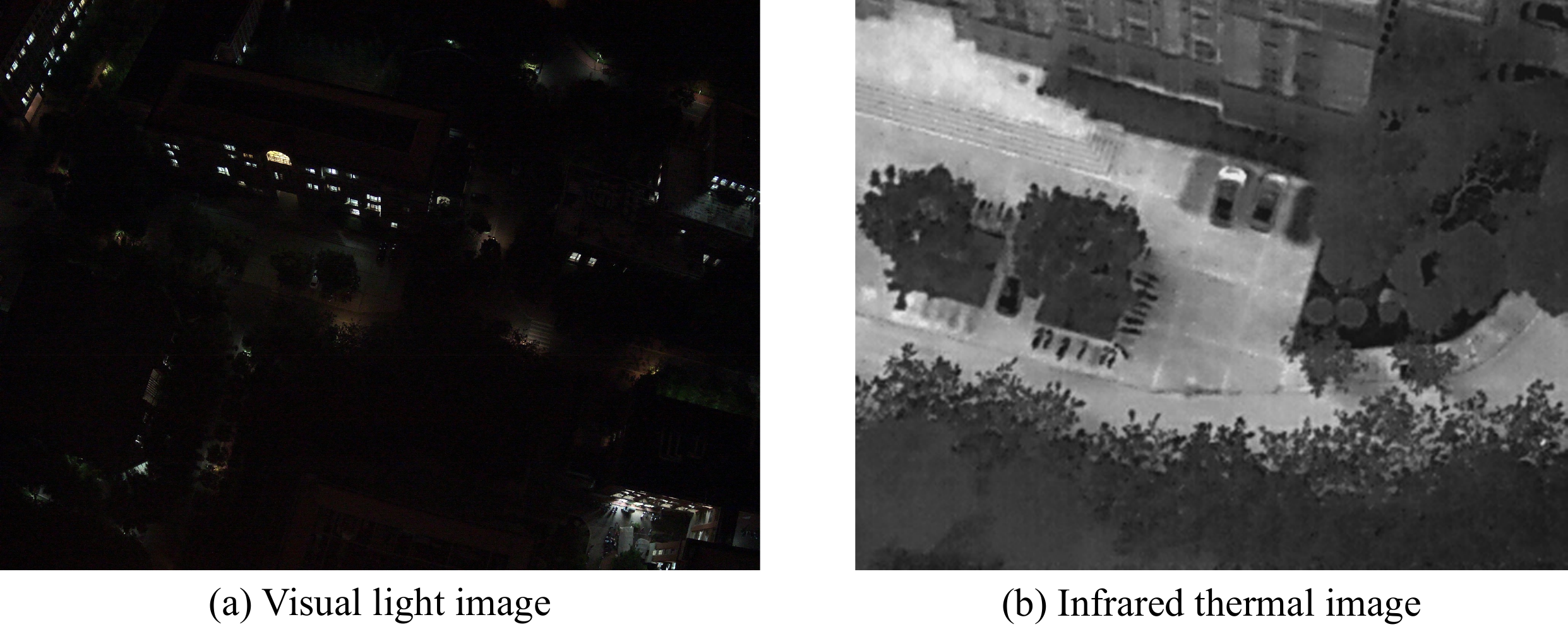}
\caption{The sample images captured by visual light and infrared thermal cameras under the same flight altitude and camera angle at night. 
The infrared thermal image readily identifies car and bicycle objects, while the visual light image faces difficulty in doing so.
The results demonstrate the superior performance of infrared thermal cameras in enabling UAVs to perform tasks more effectively during nighttime operations.}
\label{sample_different_camera}
\end{figure}

\begin{figure}[!ht]
\centering
\includegraphics[width=2.8in]{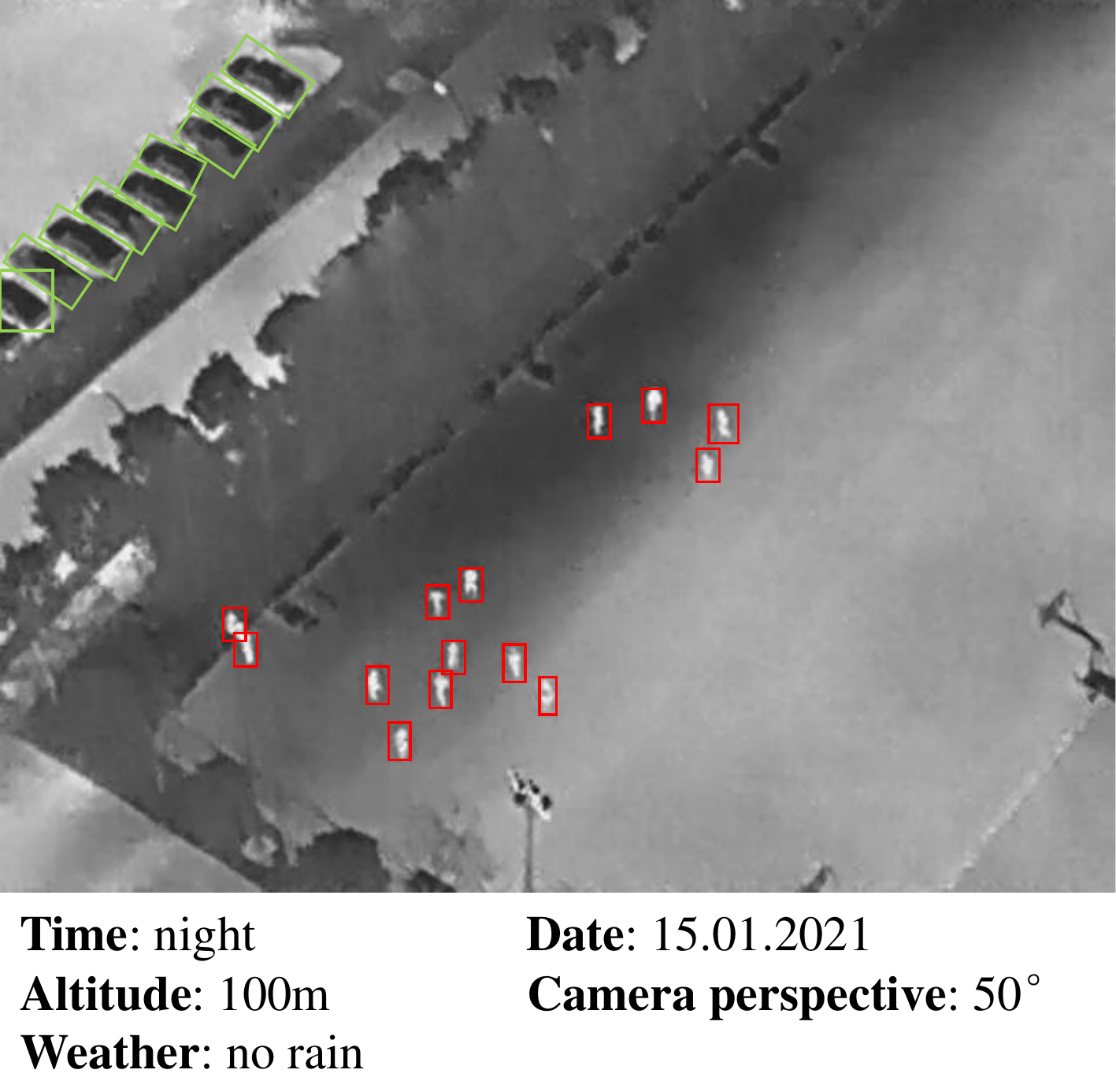}
\caption{A sample image and recorded information of the HIT-UAV.}
\label{sample_image_HIT_UAV}
\end{figure}

\begin{figure}[!ht]
\centering
\includegraphics[width=6.0in]{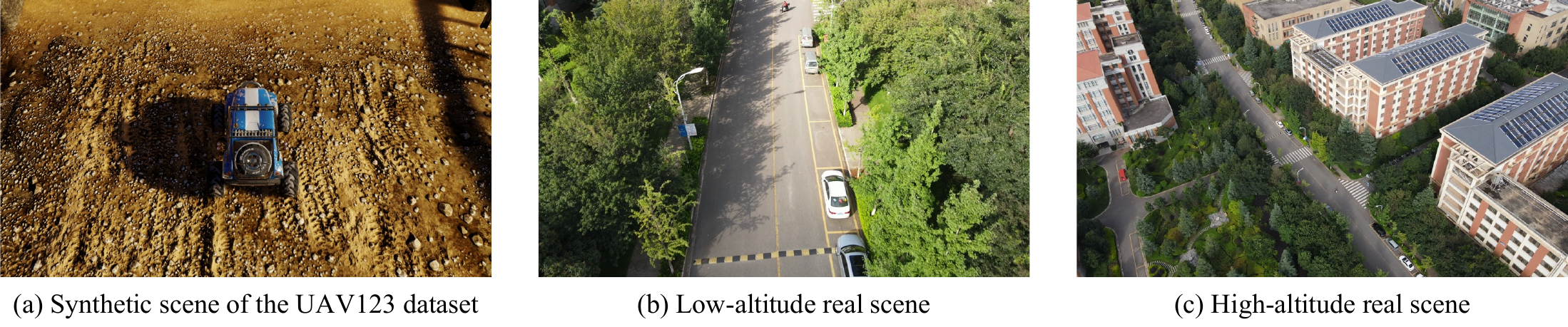}
\caption{The sample images from synthetic scenes, low-altitude real scenes, and high-altitude real scenes. Synthetic scenes often lack the lighting variations and details present in real scenes, which can result in poorer detection performance when models trained on synthetic scenes are applied to real scenes. Compared to low-altitude perspectives, high-altitude perspectives can detect more objects and enable UAVs to scan a larger area. Additionally, flying at higher altitudes allows UAVs to access areas with tall buildings, making high-altitude datasets advantageous for practical tasks. These advantages highlight the importance of high-altitude datasets in expanding the application of UAVs in real-world scenarios.}
\label{sample_different_scene}
\end{figure}

However, although many datasets have been introduced for object detection on UAVs, there are many challenges in this field:
\begin{itemize}
    \item \textit{Limited application range.} 
    Several extant UAV-based datasets only comprise visual light images, which limits their use during night-time operations and raises privacy concerns.
    As shown in Figure \ref{sample_different_camera}, infrared thermal cameras offer distinct advantages over visual light cameras for night-time imaging. 
    Additionally, Figure \ref{sample_image_HIT_UAV} shows a sample image from the HIT-UAV~\cite{HIT-UAV}, wherein persons are represented as white blocks devoid of any personal appearance, clothing, or gender information, thus ensuring complete protection of individual privacy.
    \item \textit{Insufficient record information.}
    Numerous UAV-based datasets lack critical flight information, such as altitude and camera perspective, thereby precluding researchers from investigating pertinent issues, such as the influence of these factors on detection accuracy.
    Table \ref{table_UAV_dataset_comparison} shows the record information of different datasets.
    \item \textit{Non-diversified data distribution.}
    Many UAV-based datasets focus on a narrow range of aspects, such as synthetic scenes~\cite{mueller2016benchmark, bondi2020birdsai}, low altitudes~\cite{mueller2016benchmark, hsieh2017drone, portmann2014people, li2020object}, single scenes~\cite{robicquet2016learning, portmann2014people}, or specific object categories~\cite{hsieh2017drone, shamsoshoara2021aerial, dronergbt, dronevehicle}. 
    The limitations of synthetic scenes and low altitudes are highlighted in Figure \ref{sample_different_scene}, which illustrates their drawbacks using sample images.
    Moreover, focusing on a single scene or object category restricts the applicability of the datasets in various scenarios, such as object detection in multiple scenes and detecting multiple object categories. 
    To provide a comprehensive understanding of the current UAV-based infrared thermal datasets and their drawbacks, Figure \ref{table2_sample_different_dataset} (c) - (h) are presented.
\end{itemize}

\begin{table}[!ht]
\centering
\caption{The record information of different datasets.}
\scalebox{0.8}{
\begin{tabular}{c|cccccc}
\toprule
\textbf{Dataset} & \textbf{Data type} & \textbf{Object annotation} & \textbf{Visual data} & \textbf{Altitude} & \textbf{Camera perspective} & \textbf{Infrared thermal} \\
\midrule
Stanford~\cite{robicquet2016learning} & real & yes & yes & no & no & no \\
UAV123~\cite{mueller2016benchmark} & synthetic/real & yes & yes & no & no & no \\
CARPK~\cite{hsieh2017drone} & real & yes & yes & no & no & no \\
VisDrone~\cite{zhu2018vision} & real & yes & yes & no & no & no \\
AU-AIR~\cite{bozcan2020air} & real & yes & yes & yes & no & no \\
ASL-TID~\cite{portmann2014people} & real & yes & yes & no & no & yes \\
BIRDSAI~\cite{bondi2020birdsai} & synthetic/real & yes & yes & no & no & yes \\
FLAME~\cite{shamsoshoara2021aerial} & real & no & yes & no & no & yes \\
DroneRBGT~\cite{dronergbt} & real & yes & yes & no & no & yes \\
DroneVehicle~\cite{dronevehicle} & real & yes & yes & yes & yes & yes \\
Salient Map~\cite{li2020object} & real & yes & yes & no & no & yes \\
\midrule
\textbf{HIT-UAV}~\cite{HIT-UAV} & \textbf{real} & \textbf{yes} & \textbf{yes} & \textbf{yes} & \textbf{yes} & \textbf{yes}
\\
\bottomrule
\end{tabular}}
\label{table_UAV_dataset_comparison}
\end{table}

To overcome the aforementioned challenges, we present the HIT-UAV~\cite{HIT-UAV} dataset.
The HIT-UAV~\cite{HIT-UAV} comprises infrared thermal images collected to expand the \textit{application range} of UAVs at night.
To facilitate research on diverse issues, such as the impact of UAV flight altitude and camera perspective on object detection accuracy, the HIT-UAV~\cite{HIT-UAV} records crucial \textit{information}, including flight altitude, camera perspective, daylight intensity, and image shooting date.
Figure \ref{sample_image_HIT_UAV} shows a sample image and the recorded information of the HIT-UAV~\cite{HIT-UAV}.
Covering a wide range of aspects, including higher altitudes (ranging from 60 to 130 meters), different camera perspectives (ranging from 30 to 90 degrees), various scenes (such as schools, parking lots, roads, and playgrounds), and different common object categories (such as persons, cars, bicycles, and vehicles), the HIT-UAV~\cite{HIT-UAV} aims to increase \textit{data distribution} for various tasks. 

The dataset comprises 2898 infrared thermal images extracted from 43470 frames in hundreds of videos, and all frames were collected in public and desensitized.
To promote effective use of the dataset on different tasks, the HIT-UAV~\cite{HIT-UAV} provides two types of annotated bounding boxes for each object in the images: oriented and standard.
The oriented bounding box solves the issue of significant overlap between object instances in aerial images, while the standard bounding box facilitates efficient use of the dataset. 
The HIT-UAV~\cite{HIT-UAV} includes five object categories, namely \textit{Person}, \textit{Car}, \textit{Bicycle}, \textit{OtherVehicle}, and \textit{DontCare}, with a total of 24899 annotated objects.
The \textit{DontCare} category includes objects that could not be accurately categorized by the annotators (as further detailed in the Methods section).
The dataset comprises 2029 training images, 579 test images, and 290 validation images.
To evaluate the HIT-UAV~{\cite{HIT-UAV}}, we trained and tested the well-established object detection algorithms, namely YOLOv4~{\cite{bochkovskiy2020yolov4}}, YOLOv4-tiny, Faster R-CNN~{\cite{faster-rcnn}}, and SSD~{\cite{ssd}}, using the dataset. 
The results show that compared to other visual light datasets, the algorithms exhibit exceptional performance on the HIT-UAV~{\cite{HIT-UAV}}, indicating the potential of infrared thermal datasets to improve object detection applications in UAVs significantly.
Further, we conducted an analysis of the performance of YOLOv4 and YOLOv4-tiny at different altitudes and camera perspectives, yielding insightful observations to aid users in their understanding of UAV-based object detection.

\textbf{To the best of our knowledge, the HIT-UAV~\cite{HIT-UAV} is the first publicly available high-altitude UAV-based infrared thermal dataset for detecting persons and vehicles.
The HIT-UAV~\cite{HIT-UAV} has the great potential to enable several research activities,} such as (1) the application range of infrared thermal cameras in object detection tasks, (2) the feasibility of UAV-based search and rescue missions at night, (3) the relationship of flight altitude and object detection precision on UAVs, (4) the impact of camera perspective for UAV-based object detection.

\begin{table}[ht]
\centering
\caption{DJI Matrice M210 setup.}
\scalebox{0.8}{
\begin{tabular}{cc}
\toprule
\textbf{Dimensions} & \begin{tabular}[c]{@{}c@{}}Unfolded, 883×886×398 mm;\\      Folded, 722×282×242 mm\end{tabular} \\
\midrule
\textbf{Diagonal Wheelbase} & 643 mm \\
\midrule
\textbf{Weight} & \begin{tabular}[c]{@{}c@{}}Approx. 4.8 kg\\ (with two TB55 batteries) \end{tabular} \\
\midrule
\textbf{Max Takeoff Weight} & 6.14 kg \\
\midrule
\textbf{Max Payload} & 1.34 kg \\
\midrule
\textbf{Max Angular Velocity} & Pitch: 300°/s,   Yaw: 120°/s \\
\midrule
\textbf{Max Ascent Speed} & 16.4 ft/s (5   m/s) \\
\midrule
\begin{tabular}[c]{@{}c@{}}\textbf{Max Descent Speed} \\ \textbf{(vertical)}\end{tabular} & 9.8 ft/s (3   m/s) \\
\midrule
\textbf{Max Speed} & \begin{tabular}[c]{@{}c@{}}S-mode/A-mode: 73.8 kph (45.9 mph);\\      P-mode: 61.2 kph (38 mph)\end{tabular} \\
\midrule
\begin{tabular}[c]{@{}c@{}}\textbf{Max Flight Time} \\ \textbf{(with two TB55 batteries)}\end{tabular} & \begin{tabular}[c]{@{}c@{}}34 min (no payload);\\      24 min (takeoff weight: 6.14 kg)\end{tabular}
\\
\bottomrule
\end{tabular}}
\label{table3_M210_setup}
\end{table}

%% file: 3_Method.tex
The UAV platform selected for image capture was the DJI Matrice M210 V2~\cite{DJIM210V2}, which costs approximately 10,000 US dollars.
The setup of the DJI Matrice M210 V2 used is detailed in Table \ref{table3_M210_setup}.
The DJI Zenmuse XT2 camera~\cite{DJIXT2} was loaded on the UAV to capture the images. 
The DJI Zenmuse XT2 camera features a FLIR longwave infrared thermal camera with a thermal infrared camera resolution of 640$\times$512 pixels and a 25mm lens, as well as a visual camera that captures 4K videos and 12MP photos. 
The cost of the DJI Zenmuse XT2 camera is approximately 8000 US dollars. 

The dataset generation pipeline comprise four stages: video capture, frame extraction and data cleaning, object annotation, and dataset generation.

\subsection*{Video capture} 
We captured videos under varying conditions, including schools, parking lots, roads, playgrounds, and more.
The flight altitude ranged from 60 to 130 meters, and the camera perspective ranged from 30 to 90 degrees.
We conducted flights during both day and night time.
For each video, we recorded the flight altitude, camera perspective, flight date, and daylight intensity.

\subsection*{Frame extraction and data cleaning}
There is a slight variation in image features between consecutive video frames, making most frames unsuitable for improving the performance of object detection model. 
Although many datasets reserve full frames to train detection models, this approach does not address the limited feature distribution problem. Fortunately, the HIT-UAV~\cite{HIT-UAV} provides a sufficient number of original frames (43470 frames) to ensure a wide distribution of features.
The frame resolution is 640$\times$512, bit depth is 8, and the average compression rate is 21.059\%.
To filter adjacent frames that have little difference, we sampled an image every 15 frames (since the video refresh rate is 7 FPS), resulting in 2898 infrared thermal images.

\begin{figure}[!ht]
\centering
\includegraphics[width=6.2in]{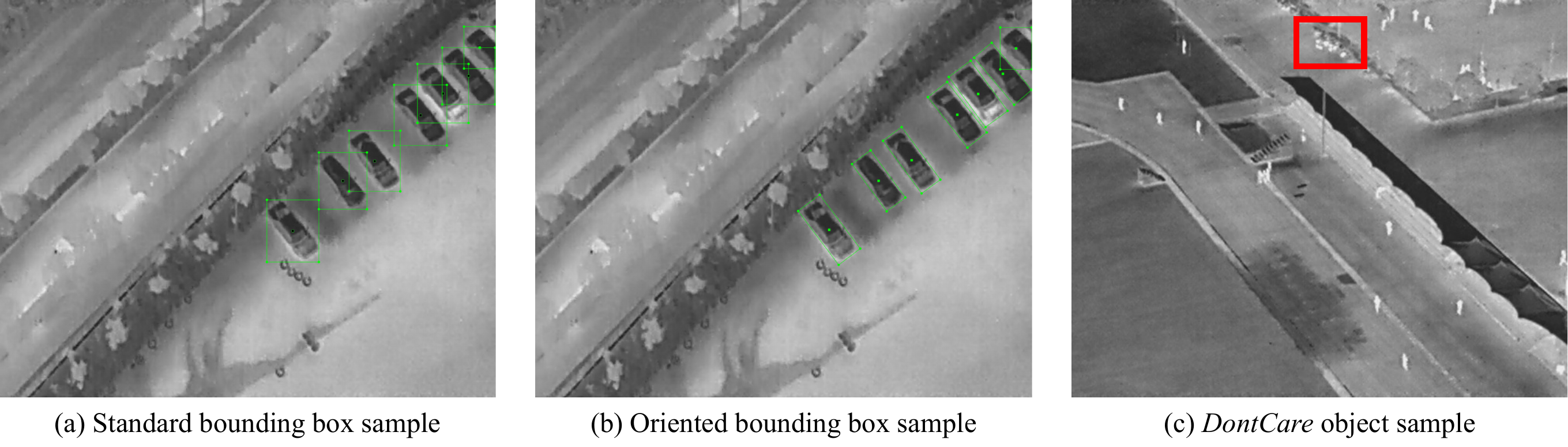}
\caption{The samples of the standard bounding box, oriented bounding box, and \textit{DontCare} object. 
Oriented bounding boxes have a smaller overlap than standard bounding boxes.
In the (c), the red box represents the \textit{DontCare} object.
It is difficult to accurately identify whether the objects in this area are people or not.}
\label{box_vis}
\end{figure}

\subsection*{Object annotation}
We annotated the objects in the dataset using two types of bounding boxes: standard and oriented.
The standard bounding box is represented as $(x_c, y_c, w, h)$, where $(x_c, y_c)$ denotes the center coordinate and $w$ and $h$ denote the width and height of the bounding box, respectively.
However, accurately labeling objects in aerial images from the perspective of UAVs can be challenging.
To address this issue, we used $\theta$-based oriented bounding box~\cite{yao2012detecting} to label object instances.
The oriented bounding box is represented as $(x_c, y_c, w, h, \theta)$, where $\theta$ denotes the oriented angle from the horizontal direction of the standard bounding box.
As shown in Figure \ref{box_vis} (a), the overlap of standard bounding boxes can be significant, making it difficult for state-of-the-art object detection algorithms to distinguish them well.
Using oriented bounding boxes accurately annotates the objects and solves this issue, as shown in Figure \ref{box_vis} (b).
Note that the bounding box on the boundary is standard because the oriented bounding box cannot exceed the edge.
One drawback of oriented bounding boxes is that few native object detection algorithms support training with them.
To help users utilize the dataset, we provide both oriented and standard bounding box annotation files.

We performed manual annotation of oriented object bounding boxes for all images using a modified version of the LabelImg tool.
Difficult and truncated object instances were also labeled.
Three individuals were involved in the annotation process, and each annotation was verified by the others.
To facilitate the use of the dataset, we developed a tool to convert oriented bounding boxes to standard bounding boxes.
The conversion method is as follows: First, we obtained the minimum and maximum x and y coordinates $(x_{min}, x_{max}, y_{min}, y_{max})$ of the oriented bounding box.
Then, we used $(x_{min}, x_{max}, y_{min}, y_{max})$ as the boundary to obtain the standard bounding box, where the center coordinate was calculated as $x_c = (x_{min} + x_{max}) / 2$ and $y_c = (y_{min} + y_{max}) / 2$, and the width and height were calculated as $w = x_{max} - x_{min}$ and $h = y_{max} - y_{min}$, respectively.

\subsection*{Dataset generation}
We developed a dataset generation tool with functions that include XML and JSON label file generation and dataset splitting.
The original images were organized into different folders based on flight data, and the tool generated XML and JSON label files corresponding to each image.
To facilitate object detection model training, we split the dataset into training, test, and validation sets with a ratio of 70\%, 20\%, and 10\%, respectively, using the Hold-out method~\cite{holdout}.

\begin{figure}[!ht]
\centering
\includegraphics[width=6.2in]{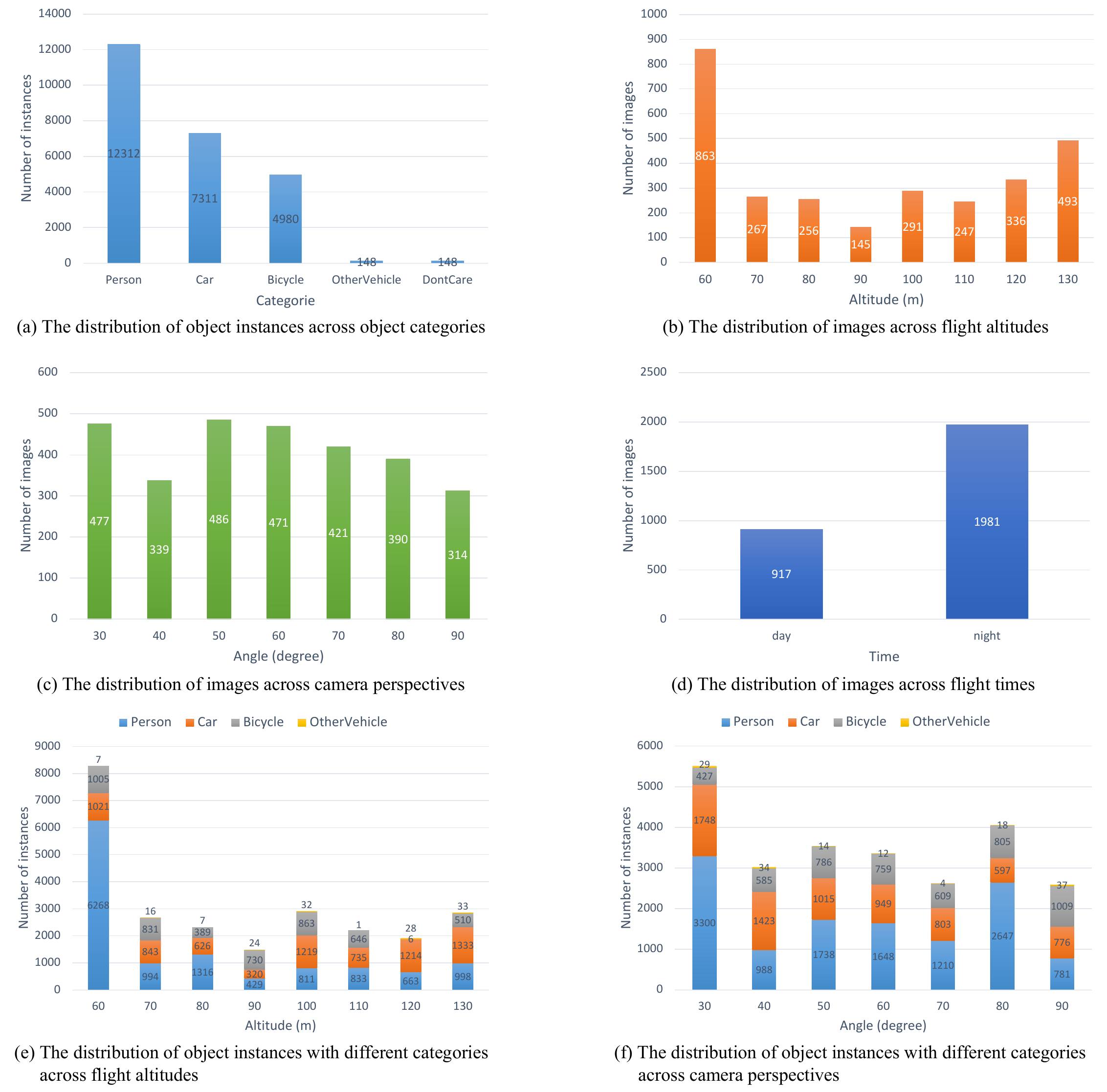}
\caption{The data distribution of the HIT-UAV.}
\label{dataset_distribution}
\end{figure}

%% file: 4_Data_records.tex
The dataset is available at Zenodo~\cite{HIT-UAV}.

\subsection*{Folder structure and recording format}
We offer two types of annotation files for users: XML files based on the VOC dataset format and JSON files based on the MS COCO dataset format.
Both of these formats are commonly used benchmarks for object detection in computer vision.
The top-level folder of our dataset includes four subfolders: \textit{normal\_json}, \textit{normal\_xml}, \textit{rotate\_json}, and \textit{rotate\_xml}.
The \textit{normal\_json} and \textit{normal\_xml} folders contain annotation files with standard bounding boxes in JSON and XML formats, respectively.
On the other hand, the \textit{rotate\_json} and \textit{rotate\_xml} folders contain annotation files with oriented bounding boxes in JSON and XML formats, respectively.

The image files are named according to the following format: $T\_HH(H)\_AA\_W\_NNNNN$, where $T$ indicates the shooting time (0 for day, 1 for night), $HH(H)$ indicates the flight altitude (ranging from 60 to 130 meters), $AA$ denotes the camera perspective (ranging from 30 to 90 degrees), $W$ indicates the weather condition (only images captured under no rain conditions were included in the dataset), and $NNNNN$ denotes the serial number of the image.

\subsection*{Properties}
The annotated object categories include four types that highly appear in rescue and search missions: \textit{Person}, \textit{Car}, \textit{Bicycle}, \textit{OtherVehicle}.
In addition, we labeled unrecognizable objects, namely \textit{DontCare}, because many objects cannot identify specific types by annotator in high aerial images.
As shown in Figure \ref{box_vis} (c), the red box represents the object of \textit{DontCare}.
In this object area, it is difficult to accurately identify if they are persons.
Therefore, the \textit{DontCare} can point out easily confused objects in the image.

\begin{figure}[!ht]
\centering
\includegraphics[width=5.0in]{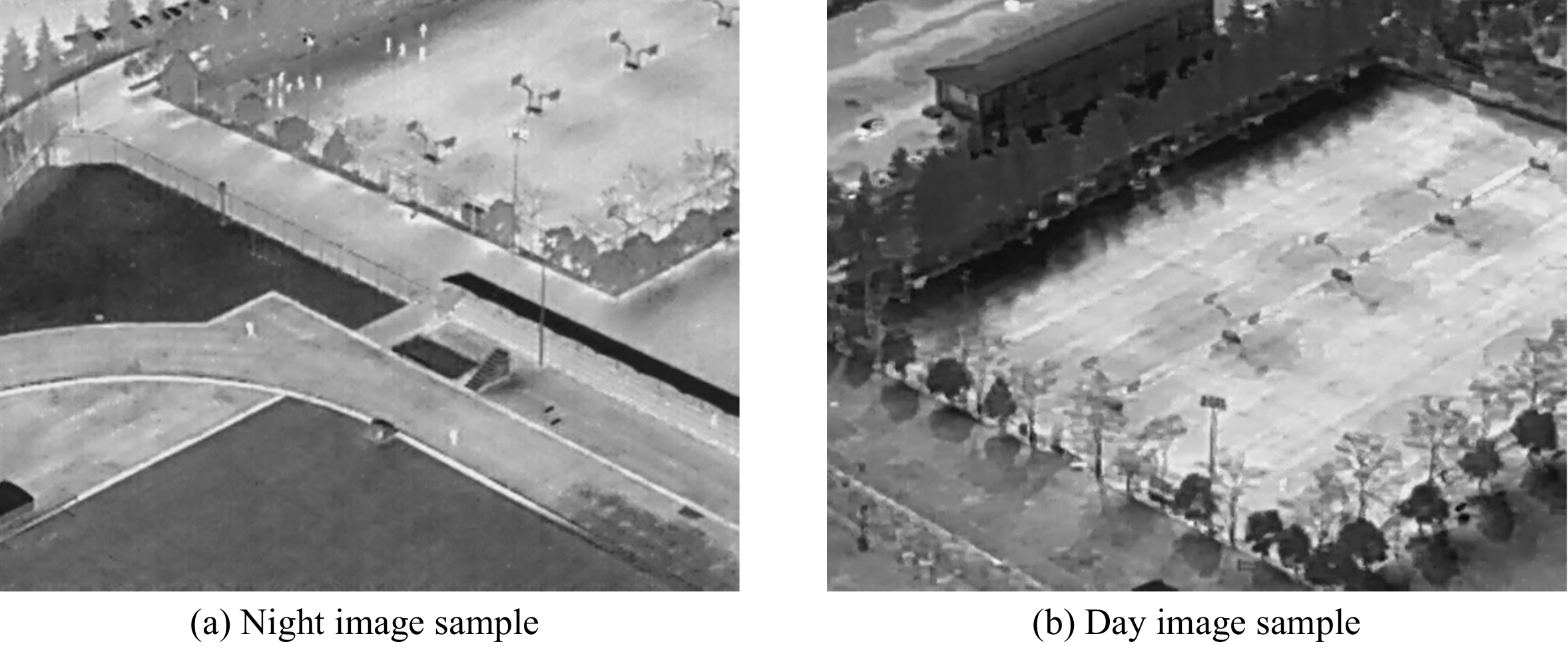}
\caption{The samples of the night and day images.}
\label{night_day_vis}
\end{figure}

\begin{figure}[!ht]
\centering
\includegraphics[width=6.5in]{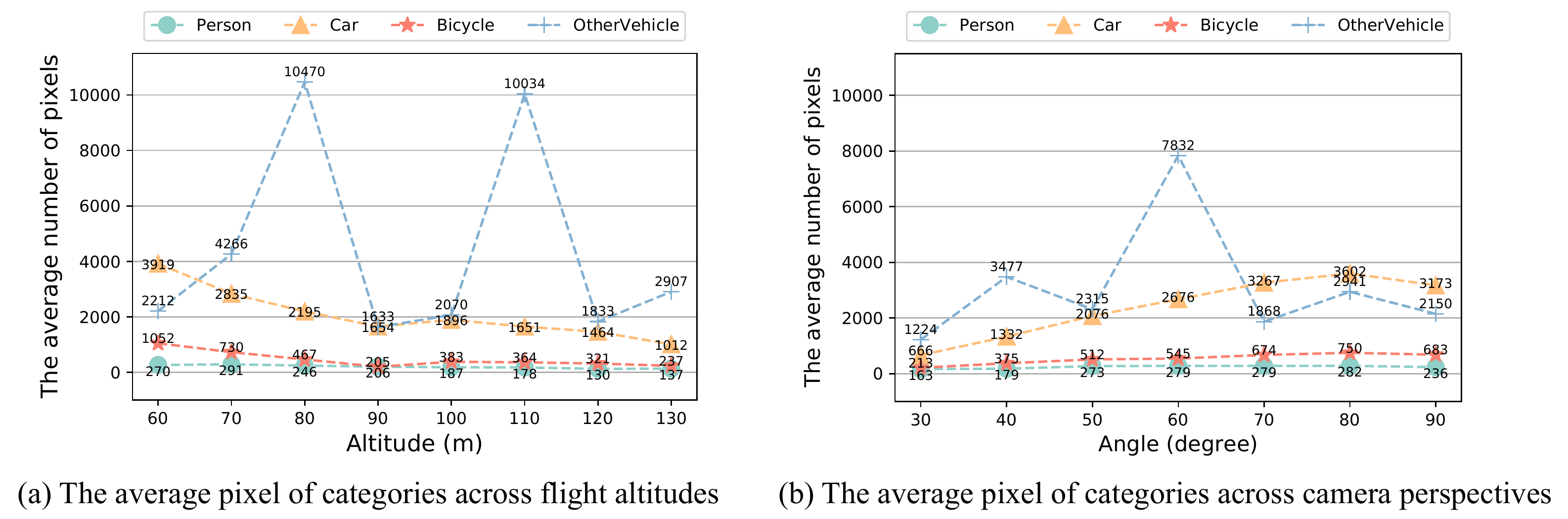}
\caption{The average pixel of categories across flight altitudes and camera perspectives.}
\label{dataset_pixel}
\end{figure}

\begin{figure}[!ht]
\centering
\includegraphics[width=6.0in]{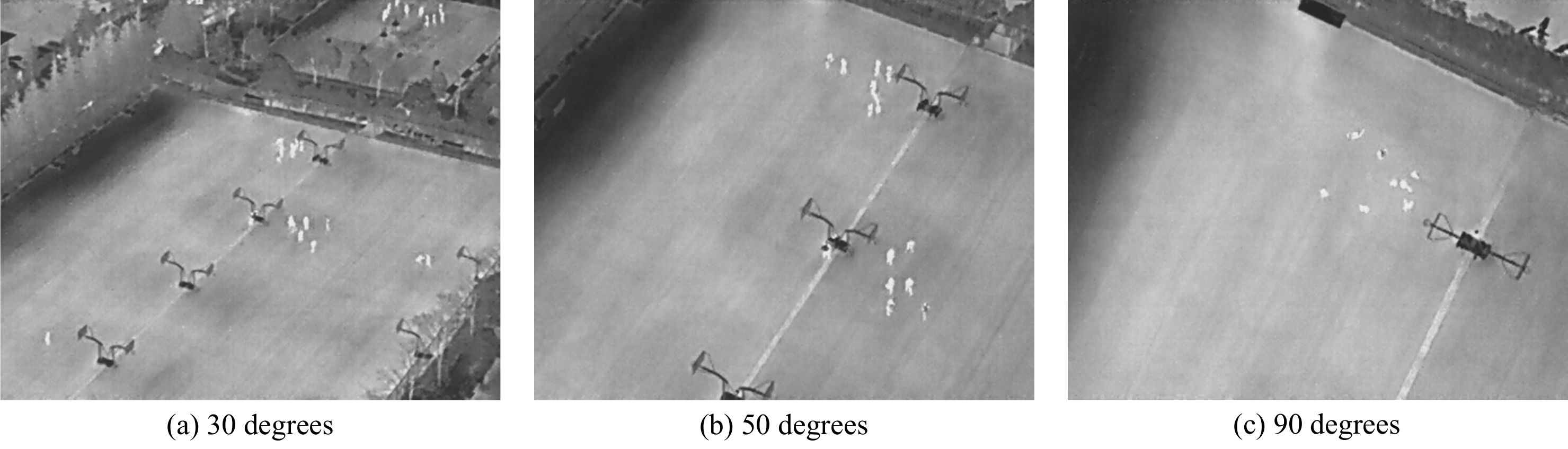}
\caption{The sample images taken at 80 meters with varying camera perspectives.
At 30 degrees, objects in the far distance appear smaller due to the wider field of view.
Conversely, at 50 degrees, objects appear larger.
However, at 90 degrees, objects once again become smaller due to the reduction in the visible surface area of objects.}
\label{fig_angle_comp}
\end{figure}

Figure \ref{dataset_distribution} (a) shows the distribution of annotations across object categories.
The main object for the rescue mission (\textit{Person}) appears more than other objects.
Additionally, the presence of a substantial number of \textit{Car} and \textit{Bicycle} objects makes the HIT-UAV~\cite{HIT-UAV} suitable for a wide range of common tasks.
To enhance the versatility of the dataset for high-altitude missions, flight altitudes were recorded in intervals of 10 meters, ranging from 60 to 130 meters.
This information is depicted in Figure \ref{dataset_distribution} (b).
The camera perspectives were also recorded in increments of 10 degrees, varying from 30 to 90 degrees, as shown in Figure \ref{dataset_distribution} (c).
Infrared thermal images have a significant difference between day and night due to the higher background temperature during the day.
As shown in Figure \ref{night_day_vis}, the infrared thermal image during the night is easier to identify the objects than during the day because the background temperature of the night is lower than the day.
To increase the diversity of the dataset, infrared thermal images were collected both during the day and night, as presented in Figure \ref{dataset_distribution} (d).
Figure \ref{dataset_distribution} (e) and (f) present the distribution of instances with varying categories across flight altitudes and camera perspectives, respectively.
The average pixels of different categories across flight altitudes are depicted in Figure {\ref{dataset_pixel}} (a). 
Theoretically, the average pixel size is expected to decrease with increasing altitude, since higher altitudes result in smaller object sizes. 
However, for the category of \textit{OtherVehicle}, the fluctuations are large due to its limited number of instances, which leads to the influence of truncated objects on the results. 
The average pixel of the remaining categories generally decreases with altitude, though there may be slight fluctuations due to the difference in image coverage at different altitudes and angles. 
The average pixel of categories across camera perspectives is shown in Figure {\ref{dataset_pixel}} (b), where it is observed that the average pixel size increases initially and then decreases with increasing angle. 
This is due to the fact that objects become more prominent with reduced vision, but their visible surface area decreases with greater angles.
Figure {\ref{fig_angle_comp}} illustrates these visual changes.

%% file: 5_Technical_validation.tex
We trained four well-established object detection algorithms, namely YOLOv4, YOLOv4-tiny, Faster-RCNN, and SSD, using the HIT-UAV{~\cite{HIT-UAV}}.
The dataset consisted of 2029 training images, 290 validation images, and 579 test images.
The experiments were performed on an RTX 2080Ti GPU.
YOLOv4 and YOLOv4-tiny were trained using the Darknet framework, while Faster-RCNN (with a ResNet-101 backbone) and SSD-512 were trained using the MMDetection~{\cite{mmdetection}} framework.
The pre-trained models for YOLOv4 and YOLOv4-tiny were obtained from official sources.
The training process was performed for a maximum of 10,000 steps, with a batch size of 64 and subdivision of 16. 
The learning rate was set to 0.0013 and was multiplied by 0.1 at steps 8000 and 9000. The weight decay and momentum were set to 0.949 and 0.0005. 
For Faster-RCNN and SSD, the official ResNet-101 and VGG16 models were used as pre-trained models. 
The maximum number of epochs was 32, with a batch size of 16. 
The learning rate was set to 0.02 and had a warm-up ratio of 0.001, with a warm-up iteration of 500. 
The weight decay and momentum were set to 0.9 and 0.0001.

Table \ref{table4_precision_baseline_model} presents the precision of the aforementioned models on the HIT-UAV{~\cite{HIT-UAV}} test set, as well as the precision of YOLOv4 and YOLOv4-tiny trained on the COCO dataset and the highest accuracy (attained by RRNet) on the VisDrone-2019 challenge~\cite{zhu2020vision}.
Our observations indicate that the Average Precision (AP) value for the category of \textit{Person} is significantly lower when using YOLOv4-tiny on the HIT-UAV~\cite{HIT-UAV}. This discrepancy may be attributed to the lower detection capability of YOLOv4-tiny for small objects in comparison to other models.
Additionally, the AP for the category of \textit{OtherVehicle} is subpar, which may be due to the category imbalance issue. 
The SSD-512 model exhibits improved performance in the imbalanced category.
In the VisDrone-2019 challenge, the highest precision of 55.82\% mean Average Precision (mAP) was achieved by the RRNet method.
However, the official YOLOv4 model achieved 65.7\% mAP on the COCO dataset, surpassing RRNet in the VisDrone challenge.
This indicates that aerial image information is more complex than that of natural images.
Finally, for the HIT-UAV~\cite{HIT-UAV}, YOLOv4 achieved an mAP of 84.75\%, indicating the following observations:

\begin{itemize}
    \item 
    Infrared thermal images effectively filter out extraneous information, leading to improved object identification.
    \item
    Infrared thermal images facilitate the outstanding performance of common detection models with limited image data, due to the easily recognizable features of the objects in such images.
    The HIT-UAV~\cite{HIT-UAV} has the potential to facilitate the detection of vehicles and persons by UAVs.
\end{itemize}

\begin{table}[!ht]
\centering
\caption{The Average Precision (AP) of the baseline models.}
\scalebox{0.7}{
\begin{tabular}{c|c|cccc|c}
\toprule
\textbf{Model} & \textbf{Dataset} & \textbf{\textit{Person} AP (\%)} & \textbf{\textit{Car} AP (\%)} & \textbf{\textit{Bicycle} AP (\%)} & \textbf{\textit{OtherVehicle} AP (\%)} & \textbf{mAP@0.50 (\%)} \\
\midrule
\midrule
YOLOv4 &
HIT-UAV &
  \begin{tabular}[c]{@{}c@{}}89.88\\      (TP = 2370, FP = 346)\end{tabular} &
  \begin{tabular}[c]{@{}c@{}}92.64\\      (TP = 1241, FP = 166)\end{tabular} &
  \begin{tabular}[c]{@{}c@{}}86.48\\      (TP = 696, FP = 158)\end{tabular} &
  \begin{tabular}[c]{@{}c@{}}69.99\\      (TP = 26, FP = 8)\end{tabular} &
  \begin{tabular}[c]{@{}c@{}}84.75\\      (TP = 4333, FP = 678, FN = 447)\end{tabular} \\
YOLOv4-tiny &
  HIT-UAV &
  \begin{tabular}[c]{@{}c@{}}16.86\\      (TP = 214, FP = 50)\end{tabular} &
  \begin{tabular}[c]{@{}c@{}}83.61\\      (TP = 1080, FP = 226)\end{tabular} &
  \begin{tabular}[c]{@{}c@{}}51.9\\      (TP = 398, FP = 182)\end{tabular} &
  \begin{tabular}[c]{@{}c@{}}49.17\\      (TP = 14, FP = 7)\end{tabular} &
  \begin{tabular}[c]{@{}c@{}}50.38\\      (TP = 1706, FP = 465, FN = 3074)\end{tabular}
  \\
Faster-RCNN & HIT-UAV & 75.5 & 95.6 & 86.4 & 46.8 & 76.8 \\
SSD-512 & HIT-UAV & 85.6 & 96.3 & 86.0 & 74.4 & 85.6 \\
YOLOv4 & COCO & \textbackslash{} & \textbackslash{} & \textbackslash{} & \textbackslash{} & 65.7 \\
YOLOv4-tiny & COCO & \textbackslash{} & \textbackslash{} & \textbackslash{} & \textbackslash{} & 40.2 \\
RRNet & VisDrone-2019 & \textbackslash{} & \textbackslash{} & \textbackslash{} & \textbackslash{} & 55.82 
\\
\bottomrule
\end{tabular}}
\label{table4_precision_baseline_model}
\end{table}

\begin{figure}[!ht]
\centering
\includegraphics[width=6.0in]{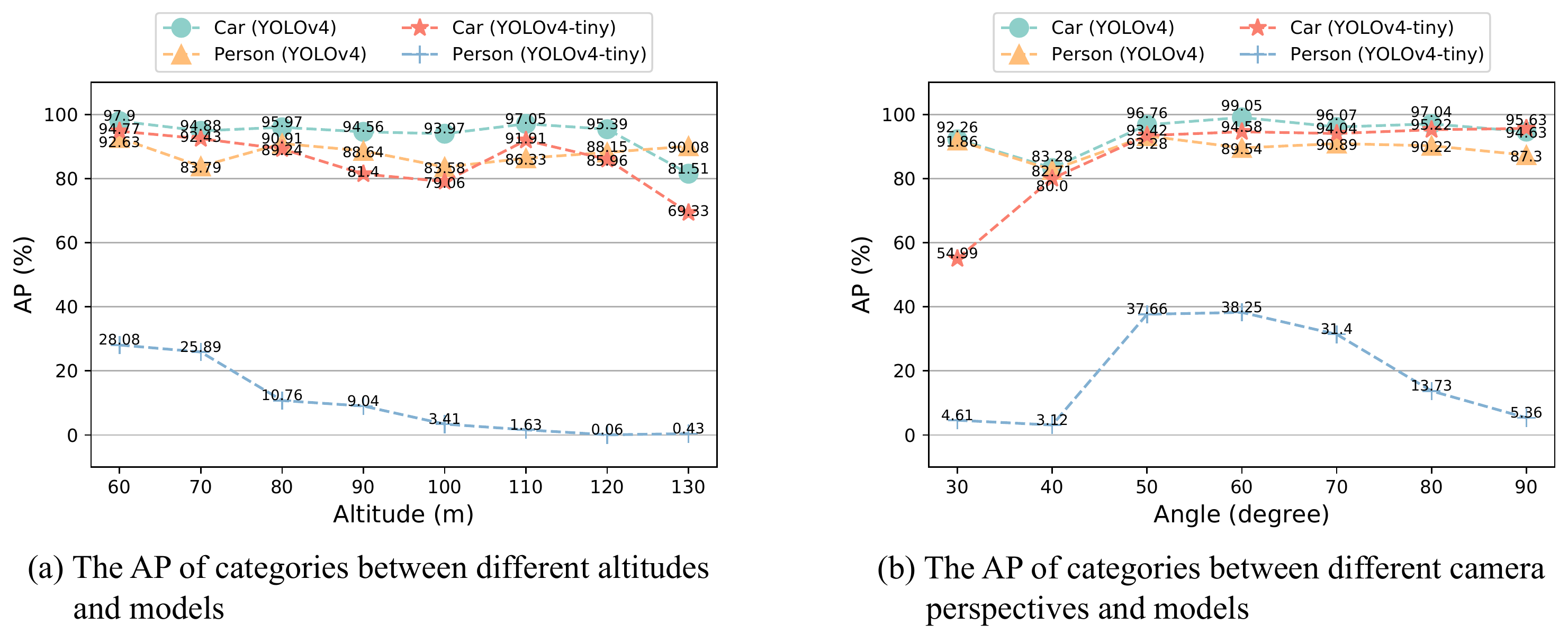}
\caption{The Average Precision (AP) of categories in the HIT-UAV test set.}
\label{fig_ap}
\end{figure}

We used YOLOv4 and YOLOv4-tiny as samples to study the relationship and impact of altitude and camera perspective on UAV-based object detection.
The categories of \textit{Person} and \textit{Car} were selected for this experiment, as the categories of \textit{OtherVehicle} and \textit{Bicycle} have a limited number of objects in the HIT-UAV{~\cite{HIT-UAV}}.
A limited number of objects would result in fluctuations in statistical results.
The results of the study are shown in Figure {\ref{fig_ap}}. 
The following observations and insights have been gleaned from the results:
\begin{itemize}
    \item 
    The AP of YOLOv4 demonstrates stability within a certain range, suggesting that variations in altitudes and angles do not significantly impact the detection performance of robust algorithms.
    \item
    The AP of YOLOv4-tiny for the \textit{Person} category tends to decrease with increasing altitude.
    This decrease is observed in three stages, ranging from 60m to 80m, 80m to 90m, and 100m to 130m, suggesting that the detection performance of lightweight algorithms is significantly impacted when objects fall outside of a certain size range.
    Higher altitudes provide a wider field of view, enabling UAVs to cover larger areas within the same flight time.
    In some UAV tasks, such as person rescue, users may need to weigh the trade-off between detection precision and altitude to achieve optimal performance.
    \item
    The AP of YOLOv4-tiny for the \textit{Person} category first increases and then decreases with increasing camera angle.
    This result highlights the impact of the visible surface of objects on detection precision.
    At 90 degrees, as shown in Figure {\ref{fig_angle_comp}} (c), individuals appear as points, making them more challenging to identify compared to when viewed at 50 degrees.
    As a result, it is crucial for users to choose the appropriate camera perspective when performing object detection tasks.
\end{itemize}

\begin{figure}[!ht]
\centering
\includegraphics[width=7in]{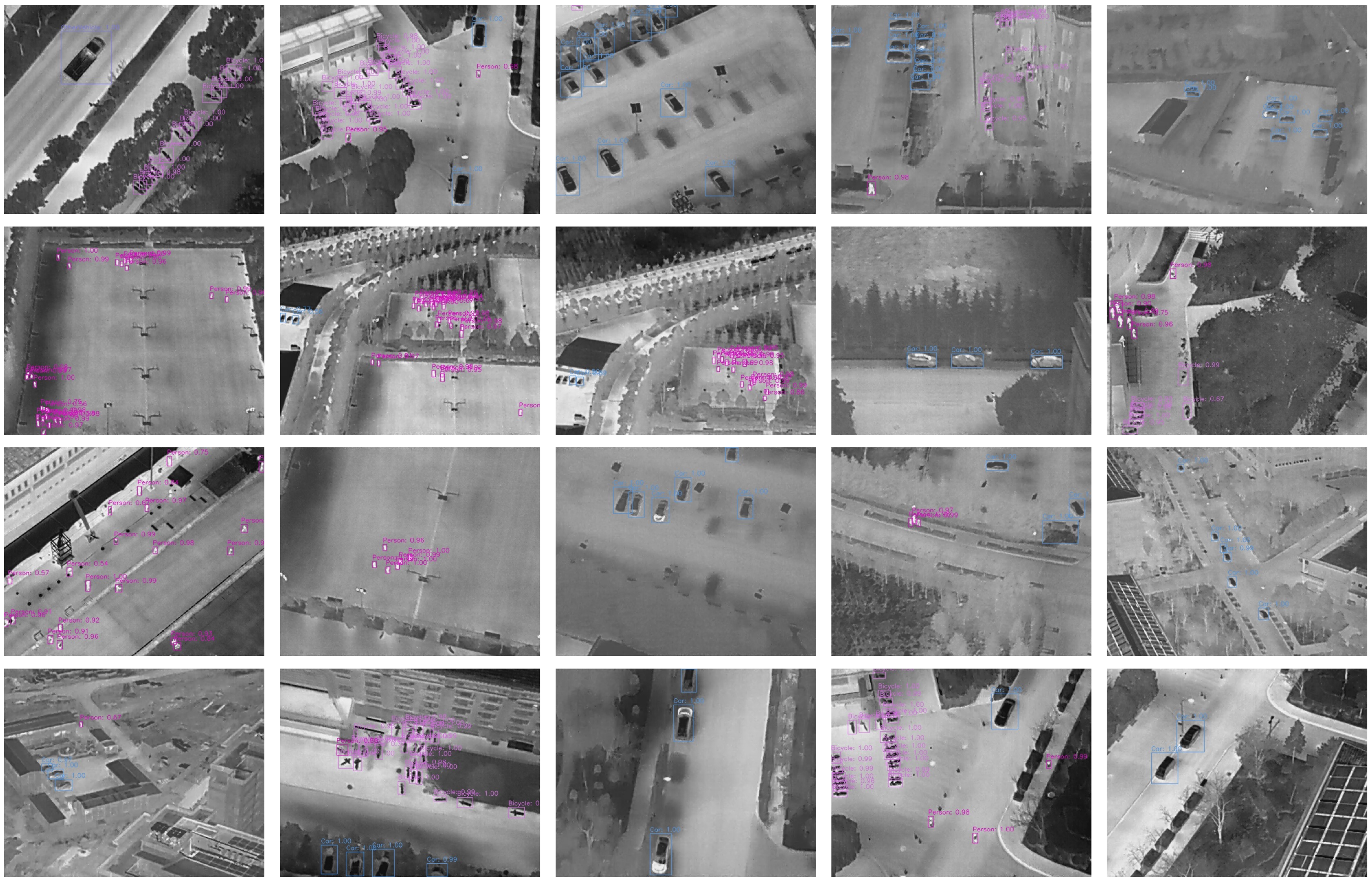}
\caption{The sample results of YOLOv4 detection.}
\label{detection_result}
\end{figure}

The sample detection results of the YOLOv4 model trained on the HIT-UAV~\cite{HIT-UAV} are shown in Figure \ref{detection_result}.
The results demonstrate that the model effectively recognizes objects in infrared thermal aerial images.
We hope the HIT-UAV~\cite{HIT-UAV} can promote the development of drone-based object detection tasks.

%% file: 6_Uasge_notes.tex
The HIT-UAV~\cite{HIT-UAV} is available at \href{https://github.com/suojiashun/HIT-UAV-Infrared-Thermal-Dataset}{https://github.com/suojiashun/HIT-UAV-Infrared-Thermal-Dataset}.
Users can download the dataset to train object detection algorithms.
The VOC and MS COCO dataset is a widely used benchmarks for object detection.
We provide the label files with VOC and MS COCO format.
Users can easily use the HIT-UAV~\cite{HIT-UAV}.

The HIT-UAV{~\cite{HIT-UAV}} was collected in a diverse range of environments, including schools, parking lots, roads, and playgrounds.
This allows for the application of trained object detection models to these scenarios as well as other environments through the generalization capabilities of deep learning.
Researchers can use the HIT-UAV~\cite{HIT-UAV} to train object detection models to research the application range of infrared thermal in different object detection tasks.
Additionally, the trained models have the potential to be employed in UAV-based search and rescue missions during nighttime to evaluate their feasibility.

%% file: 7_code_availability.tex
The data processing code is available in the \textit{tools} folder of \href{https://github.com/suojiashun/HIT-UAV-Infrared-Thermal-Dataset}{https://github.com/suojiashun/HIT-UAV-Infrared-Thermal-Dataset}.
The code is written in Python.
The functions of the tools are as follows: 
(1) The \textit{tools/devtoolkit/labelTransformer.py} is to convert oriented bounding boxes to standard bounding boxes and generate the dataset,
(2) The \textit{tools/devtoolkit/visualization.py} is to visualize images with bounding boxes,
(3) The \textit{tools/output/voc2yolo.py} is to generate the label files with the YOLO format to help users train the YOLO, which is the representative object detection algorithm.

%% file: main.bbl
\begin{thebibliography}{99}

\bibitem{wallace2012development} Wallace, L., Lucieer, A., Watson, C. \& Turner, D. Development of a UAV-LiDAR System with Application to Forest Inventory. \textit{Remote Sens.} \textbf{4(6),} 1519-1543, \href{https://doi.org/10.3390/rs4061519}{https://doi.org/10.3390/rs4061519} (2012).

\bibitem{samad2013potential} Samad, A. M., Kamarulzaman, N., Hamdani, M. A., Mastor, T. A. \& Hashim, K. A. The potential of Unmanned Aerial Vehicle (UAV) for civilian and mapping application. \textit{2013 IEEE 3rd International Conference on System Engineering and Technology} 313-318, \href{https://doi.org/10.1109/ICSEngT.2013.6650191}{https://doi.org/10.1109/ICSEngT.2013.6650191} (2013).

\bibitem{heintz2007images} Heintz, F., Rudol, P. \& Doherty, P. From images to traffic behavior - A UAV tracking and monitoring application. \textit{2007 10th International Conference on Information Fusion} 1-8, \href{https://doi.org/10.1109/ICIF.2007.4408103}{https://doi.org/10.1109/ICIF.2007.4408103} (2007).

\bibitem{bravo2019use} Bravo, R. Z. B., Leiras, A. \& Cyrino Oliveira, F. L. The Use of UAVs in Humanitarian Relief: An Application of POMDP-Based Methodology for Finding Victims. \textit{Production and Operations Management} \textbf{28(2),} 421-440, \href{https://doi.org/10.1111/poms.12930}{https://doi.org/10.1111/poms.12930} (2019).

\bibitem{pouyanfar2018survey} Pouyanfar, S. \textit{et al}. A Survey on Deep Learning: Algorithms, Techniques, and Applications. \textit{ACM Computing Surveys (CSUR)} \textbf{51(5),} 1-36, \href{https://doi.org/10.1145/3234150}{https://doi.org/10.1145/3234150} (2018).

\bibitem{shi2016edge} Shi, W., Cao, J., Zhang, Q., Li, Y. \& Xu, L. Edge Computing: Vision and Challenges. \textit{IEEE Internet of Things Journal} \textbf{3(5),} 637-646, \href{https://doi.org/10.1109/JIOT.2016.2579198}{https://doi.org/10.1109/JIOT.2016.2579198} (2016).

\bibitem{everingham2010pascal} Everingham, M. \textit{et al}. The Pascal Visual Object Classes (VOC) Challenge. \textit{International Journal of Computer Vision} \textbf{88,} 303-338 (2010).

\bibitem{lin2014microsoft} Lin, T. \textit{et al}. Microsoft COCO: Common Objects in Context. \textit{Computer Vision – ECCV 2014: 13th European Conference} 745-755 (2014).

\bibitem{deng2009imagenet} Deng, J. \textit{et al}. ImageNet: A large-scale hierarchical image database. \textit{2009 IEEE Conference on Computer Vision and Pattern Recognition} 248-255, \href{https://doi.org/10.1109/CVPR.2009.5206848}{https://doi.org/10.1109/CVPR.2009.5206848} (2009).

\bibitem{HIT-UAV} Suo, J. HIT-UAV: A High-altitude Infrared Thermal Dataset for Unmanned Aerial Vehicles (v1.2.1). \textit{Zenodo} \href{https://doi.org/10.5281/zenodo.7633134}{https://doi.org/10.5281/zenodo.7633134} (2023).

\bibitem{robicquet2016learning} Robicquet, A., Sadeghian, A., Alahi, A. \& Savarese, S. Learning Social Etiquette: Human Trajectory Understanding In Crowded Scenes. \textit{Computer Vision - ECCV 2016: 14th European Conference} 549-565 (2016).

\bibitem{mueller2016benchmark} Mueller, M., Smith, N. \& Ghanem, B. A Benchmark and Simulator for UAV Tracking. \textit{Computer Vision – ECCV 2016: 14th European Conference} 445-461 (2016).

\bibitem{hsieh2017drone} Hsieh, M., Lin, Y. \& Hsu, W. H. Drone-Based Object Counting by Spatially Regularized Regional Proposal Network. \textit{Proceedings of the IEEE International Conference on Computer Vision (ICCV)} 4145-4153 (2017).

\bibitem{zhu2018vision} Zhu, P., Wen, L., Bian, X., Ling, H. \& Hu Q. Vision Meets Drones: A Challenge. Preprint at \href{https://arxiv.org/abs/1804.07437}{https://arxiv.org/abs/1804.07437} (2018).

\bibitem{bozcan2020air} Bozcan, I. \& Kayacan, E. AU-AIR: A Multi-modal Unmanned Aerial Vehicle Dataset for Low Altitude Traffic Surveillance. \textit{2020 IEEE International Conference on Robotics and Automation (ICRA)} 8504-8510, \href{https://doi.org/10.1109/ICRA40945.2020.9196845}{https://doi.org/10.1109/ICRA40945.2020.9196845} (2020).

\bibitem{portmann2014people} Portmann, J., Lynen, S., Chli, M. \& Siegwart, R. People detection and tracking from aerial thermal views. \textit{2014 IEEE International Conference on Robotics and Automation (ICRA)} 1794-1800, \href{https://doi.org/10.1109/ICRA.2014.6907094}{https://doi.org/10.1109/ICRA.2014.6907094} (2014).

\bibitem{bondi2020birdsai} Bondi, E. \textit{et al}. BIRDSAI: A Dataset for Detection and Tracking in Aerial Thermal Infrared Videos. \textit{Proceedings of the IEEE/CVF Winter Conference on Applications of Computer Vision} 1747-1756 (2020).

\bibitem{shamsoshoara2021aerial} Shamsoshoara, A. \textit{et al}. Aerial imagery pile burn detection using deep learning: The FLAME dataset. \textit{Computer Networks} \textbf{193,} 1008001, \href{https://doi.org/10.1016/j.comnet.2021.108001}{https://doi.org/10.1016/j.comnet.2021.108001} (2021).

\bibitem{dronergbt} Peng, T., Li, Q. \& Zhu, P. RGB-T Crowd Counting from Drone: A Benchmark and MMCCN Network. \textit{Proceedings of the Asian Conference on Computer Vision (ACCV)} (2020).

\bibitem{dronevehicle} Sun, Y., Cao, B., Zhu, P. \& Hu, Q. Drone-Based RGB-Infrared Cross-Modality Vehicle Detection Via Uncertainty-Aware Learning. \textit{IEEE Transactions on Circuits and Systems for Video Technology} \textbf{32(10),} 6700-6713, \href{https://doi.org/10.1109/TCSVT.2022.3168279}{https://doi.org/10.1109/TCSVT.2022.3168279} (2022).

\bibitem{li2020object} Li, M., Zhao, X., Li, J. \& Zhu, D. OBJECT DETECTION IN UAV-BORNE THERMAL IMAGES USING BOUNDARY-AWARE SALIENCY MAPS. \textit{The International Archives of Photogrammetry, Remote Sensing and Spatial Information Sciences} \textbf{43,} 1233-1238 (2020).

\bibitem{bochkovskiy2020yolov4} Bochkovskiy, A., Wang, C. Y. \& Liao, H. Y. M. YOLOv4: Optimal Speed and Accuracy of Object Detection. Preprint at \href{https://arxiv.org/abs/2004.10934}{https://arxiv.org/abs/2004.10934} (2020).

\bibitem{faster-rcnn} Ren, S., He, K., Girshick, R. \& Sun, J. Faster R-CNN: Towards Real-Time Object Detection with Region Proposal Networks. \textit{Advances in Neural Information Processing Systems 28 (NIPS 2015)} \textbf{28,} (2015).

\bibitem{ssd} Liu, W. \textit{et al}. SSD: Single Shot MultiBox Detector. \textit{Computer Vision – ECCV 2016: 14th European Conference} 21-37 (2016)

\bibitem{DJIM210V2} DJI. \textit{Matrice M210 V2} \href{https://www.dji.com/matrice-200-series-v2}{https://www.dji.com/matrice-200-series-v2} (2021).

\bibitem{DJIXT2} DJI. \textit{Zenmuse XT2} \href{https://www.dji.com/zenmuse-xt2}{https://www.dji.com/zenmuse-xt2} (2021)

\bibitem{yao2012detecting} Yao, C., Bai, X., Liu, W., Ma, Y. \& Tu, Z. Detecting texts of arbitrary orientations in natural images. \textit{2012 IEEE Conference on Computer Vision and Pattern Recognition} 1083-1090, \href{https://doi.org/10.1109/CVPR.2012.6247787}{https://doi.org/10.1109/CVPR.2012.6247787} (2012).

\bibitem{holdout} Lee, L. C., Liong, C. Y. \& Jemain, A. A. Validity of the best practice in splitting data for hold-out validation strategy as performed on the ink strokes in the context of forensic science. \textit{Microchemical Journal} \textbf{139,} 125-133, \href{https://doi.org/10.1016/j.microc.2018.02.009}{https://doi.org/10.1016/j.microc.2018.02.009} (2018).

\bibitem{mmdetection} Chen, K. \textit{et al}. MMDetection: Open MMLab Detection Toolbox and Benchmark. Preprint at \href{https://arxiv.org/abs/1906.07155}{https://arxiv.org/abs/1906.07155} (2019).

\bibitem{zhu2020vision} Zhu, P. \textit{et al}. Detection and Tracking Meet Drones Challenge. \textit{IEEE Transactions on Pattern Analysis and Machine Intelligence} \textbf{44,} 7380-7399, \href{https://doi.org/10.1109/TPAMI.2021.3119563}{https://doi.org/10.1109/TPAMI.2021.3119563} (2021).

\end{thebibliography}
